\documentclass[conference]{IEEEtran}
\usepackage{graphicx}
\usepackage{ctable}
\usepackage{spreadtab}
\usepackage{longtable}
\usepackage{multirow}
\usepackage{subcaption} 
\usepackage{amsmath}
\usepackage{hyperref}

\usepackage [english]{babel}
\usepackage [autostyle, english = american]{csquotes}
\MakeOuterQuote{"}

\newcommand\blfootnote[1]{%
  \begingroup
  \renewcommand\thefootnote{}\footnote{#1}%
  \addtocounter{footnote}{-1}%
  \endgroup
}
 
\begin{document}

\renewcommand\IEEEkeywordsname{Keywords}

\title{Adapting Procedural Content Generation to Player Personas Through Evolution
}

\author{
\IEEEauthorblockN{Pedro M. Fernandes}
\IEEEauthorblockA{\textit{INESC-ID and Instituto Superior}\\  
\textit{Técnico, Univ. de Lisboa}\\
Lisbon, Portugal\\
pedro.miguel.rocha.fernandes@ist.utl.pt}

\and

\IEEEauthorblockN{Jonathan Jørgensen}
\IEEEauthorblockA{\textit{Norwegian University of}\\
\textit{Science and Technology}\\
Trondheim, Norway\\
jonathan.jorgensen@ntnu.no}

\and

\IEEEauthorblockN{Niels N. T. G. Poldervaart}
\IEEEauthorblockA{\textit{Leiden University}\\
Leiden, The Netherlands\\
n.t.g.poldervaart@umail.leidenuniv.nl}
}

\maketitle

\begin{abstract}
Automatically adapting game content to players opens new doors for game development. In this paper we propose an architecture using persona agents and experience metrics, which enables evolving procedurally generated levels tailored for particular player personas. Using our game, "Grave Rave", we demonstrate that this approach successfully adapts to four rule-based persona agents over three different experience metrics. Furthermore, the adaptation is shown to be specific in nature, meaning that the levels are persona-conscious, and not just general optimizations with regard to the selected metric.
\end{abstract}

\IEEEpeerreviewmaketitle

\section{Introduction}
\label{sec:introdution}
Most \blfootnote{This version is a draft of the paper later published on the proceedings of the IEEE Symposium Series on Computational Intelligence (IEEE SSCI 2021) (978-1-7281-9048-8/21/\$31.00 ©2021 IEEE)}, if not all games, are played differently by different people. Where one player might take on an aggressive style, another might prefer a more defensive approach. Identifying these styles and assigning them to player archetypes, or \textit{personas} \cite{cooper1999inmates}, is an active topic in game research, as well as other related fields, such as human computer interface research and psychology. In this paper, we ask: "if given a set of such personas, how can we adapt game content to each one specifically?". To achieve this, the architecture presented on this paper was designed, along with the game "Grave Rave" and four rule-based player agents to represent personas. The game uses levels generated by a Procedural Content Generator \cite{pcg-book} (PCG).


The player experience envisioned by a game designer and the game content produced by a PCG may not always be aligned. This is not surprising, given how there can be virtually infinite different variations of content produced by the generator. However, if given a function of what content is desirable or not, one can perform a search in the "content space" represented by the generator, and thus achieve greater control over what is presented to the player. In this project, we do exactly that, using artificial evolution and functions for evaluating "desired" content. These functions are dependent on the interaction between the content (generated levels) and the playing persona agents, and will henceforth be called "experience metrics". Thus, the content is generated with respect to these metrics, as well as a specific playing style, represented by rule-based persona agents.

\begin{figure}[h!]
\centering
\includegraphics[width=.7\linewidth]{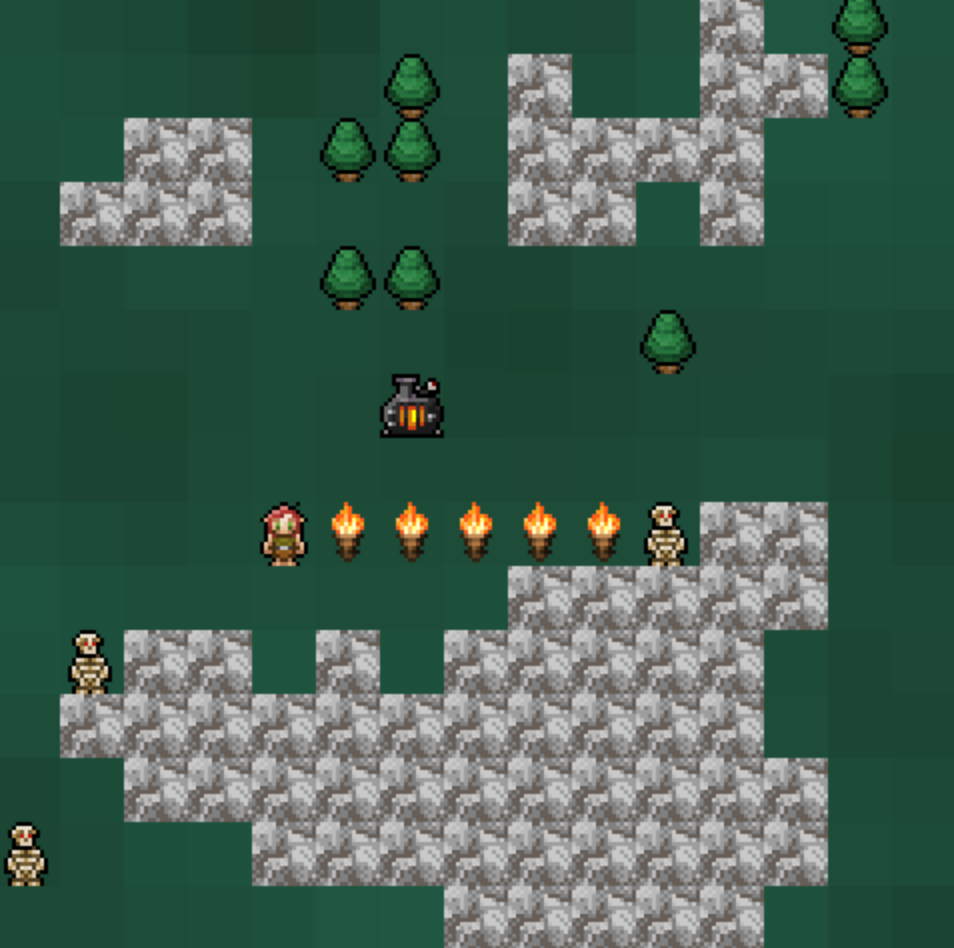}
\caption{A screenshot from "Grave Rave". The human sprite is the player character, the furnace is the base, and the skeletons are the enemies approaching the base to destroy it. The torches are projectiles thrown by the player to destroy skeletons.}
\centering
\end{figure}

Our primary contribution is an approach to generating and evaluating personalized content in games in a way that is independent of both agent architecture and game. The agent independence is achieved through using the OpenAI gym interface \cite{gym}, which is already widely adopted as a standard for reinforcement learning agents, and is technically compatible with most other agent types as well. The system is also independent on the game used, given that it accepts one or more numerical parameters to an internal procedural content generator, and that it can provide a meaningful state representation (eg. a screenshot), through the gym interface.

In our results, we show that the system quickly and consistently adapts to each persona agent. Not only with respect to the current experience metric, but also showing specific adaptation to the intended persona in comparison to the other personas. Overall, all four persona agents were tested on three metrics, namely "hardcore", "easy" and "close call", representing general and relatable experiences in games. For each of these metrics, we present an "adaptation specificity matrix", that shows the average outcome of running each agent through each of the levels generated specifically for one persona.

Our paper is organized as follows:
First, in Section \ref{sec:related}, we summarize related projects in the context of ours.
In the following section (Sec.~\ref{sec:arch}) the general architecture of the method is presented, describing each of the four main components. Then, in Section \ref{sec:impl}, we introduce the specific game, the personas and the experience metrics used in our experiments. Finally, the experimental results are presented (Sec.~\ref{sec:results}), along with discussion (Sec.~\ref{sec:discussion}), conclusion (Sec.~\ref{sec:conclusion}), and closing remarks on future work (Sec.~\ref{sec:future_work}).


\section{Related Work}
\label{sec:related}
The related works for this project spans across several central topics. Firstly, we explore PCG systems that take advantage of one or more AI agents to guide or prune level generation. Then we cover papers that address playing styles and personas, and how to represent or generate them. Finally, we explore different approaches to player adaptation and present the projects whose ideas this paper directly builds upon.

\subsection{Combining Procedural Content Generation with Agents}
Using AI agents in combination with PCG is an idea that has been explored in many different projects across the years. Volz et al. \cite{evo-mario} evolve levels for Super Mario Bros using the champion A* agent from the 2009 Mario AI competition \cite{mario-star} to evaluate the playability of their levels. Paired Open-Ended Trailblazer, or POET \cite{poet}, is an algorithm that optimize a learning agent and its environment simultaneously. Similarly, ARLPCG (Adversarial Reinforcement Learning for Procedural Content Generation) \cite{adversarial-rl} pits two RL agents, a "Generator" and a "Solver" against each other. Earle et al. \cite{learnable-generator} pose the task of content generation as a problem for a reinforcement learning agent.

\subsection{Personas and Playing Styles}

In the field of Human Computer Interaction, the concept of "persona" is often used, where the interface designers create sets of ethnographic descriptions to more easily tailor the design to certain categories of users instead of individual ones. These descriptions can be theoretical, created arbitrarily by the designer\cite{cooper1999inmates}\cite{cooper2012face} to serve as a \textit{metaphor} for a certain type of user. But personas can also be based on clustering data on the behaviour of actual users, which Canossa et al. calls \textit{lenses}\cite{play-personas}. When applied to game design, it may be useful to, for example, create a persona that represents players with a aggressive play style, while also having a persona representing those who have a more defensive approach.

Identifying and modeling different playing styles in games is not a trivial task. One interesting attempt is outlined in Holmgård et al. \cite{styles-deviation}. They use human player data from Super Mario and compare it to an agent playing the game perfectly. The different patterns of how human gameplay deviates from the optimal behaviour are clustered and categorised into distinct play-styles.

Personas that represent trends in human players are useful in many applied contexts. But the data gathering process may be resource-intensive. An alternative way is to procedurally generate players and deliberately try to find novel game play strategies. One way to do this is to generate procedural personas using MCTS \cite{automated-playtesting}. Another similar approach also uses MCTS agents, along with MAP-Elites for behavioural diversity \cite{diverse-style}. Reinforcement Learning \cite{rl} has also been used to generate personas, by changing the weight of rewards \cite{holmgaard2014evolving}. One major application of such personas is to achieve a high coverage in automated gameplay testing in commercial games.

What is demonstrated in this paper is applicable with both the more traditional definition of personas by Cooper \cite{cooper1999inmates}\cite{cooper2012face} and \textit{play-personas} as described by Conassa et al. \cite{play-personas}. In short, play-personas is a framework that extend Cooper's traditional predefined archetype personas such that personas may also be derived from clustering behaviour data of actual individual users. Play-personas start out as \textit{metaphors} with a priori attributes decided on the designer describing a certain type of user. When enough data of actual individual users is obtained, it becomes possible to add data-driven representations of user behaviours to the persona by clustering the data, also known as \textit{lenses}. In the presented architecture, both designed and data-driven personas can be used, given that they can be represented by an agent implementation.

\subsection{Adaptation to Players and Personas}
On the realm of adapting to real, particular players, González-Duque et al. \cite{fast-adapt} present "Fast Bayesian Content Adaption", where Bayesian optimization is used for dynamic difficulty adaption (DDA). Their system works online, and will after each level try to adapt it to the player, based on the time they use to solve the level, to approach a target time. Shaker et al. \cite{shaker2013fusing} also develop a system capable of generating levels that are adapted to particular players. By using behavioural cues captured from the player, they train a model to predict challenge, engagement and frustration. They further expand the work by generating Super Mario levels that maximize one of those dimensions of experience for a given particular player. 

Stepping away from directly adapting to players but instead to models of players, Togelius et al. \cite{personal-racing} evolve tracks for a racing game that maximize the "entertainment value" for modelled players.

Shu et al. \cite{experience-driven} pose endless Super Mario level generation as an RL problem, and optimize with respect to a quantified fun function, based on diversity. 

Holmgård et al. \cite{holmgaard2014personas} compare personas designed by experts and clones trained from human traces to see which ones fared better at capturing human decision making. They find that both approaches fare equally well, a result that motivates the use of hand-coded personas in this paper.

Being deeply aligned with our work, Liapis et al. \cite{liapis2015procedural} use agent personas as critics of dungeon levels. The authors define two different personas based on reward boosting, having an agent persona that prioritizes finding treasure and another that prioritizes killing monsters. They then evolve levels that maximize the utility for these two different personas, as well as evolving levels that balance utility with risk taking, ensuring the levels generated aren't completely devoid of challenge. We see our work as expanding this approach, focusing on the experience of the player and using an architecture that is theoretically valid for any type of agent and any measure of experience.

\section{Architecture}
\label{sec:arch}
Our proposed architecture for generating personalized levels based on personas has a few functional requirements. Firstly, the game used needs to have some kind of procedural generation, and the generator needs to have at least one exposed input which can influence the generated content. Secondly, there has to be an interface allowing for player agents to interact with the game. With these constraints in mind, we present the four major components of this system.

\label{sec:architecture}
\begin{figure}[h]
\includegraphics[width=\linewidth]{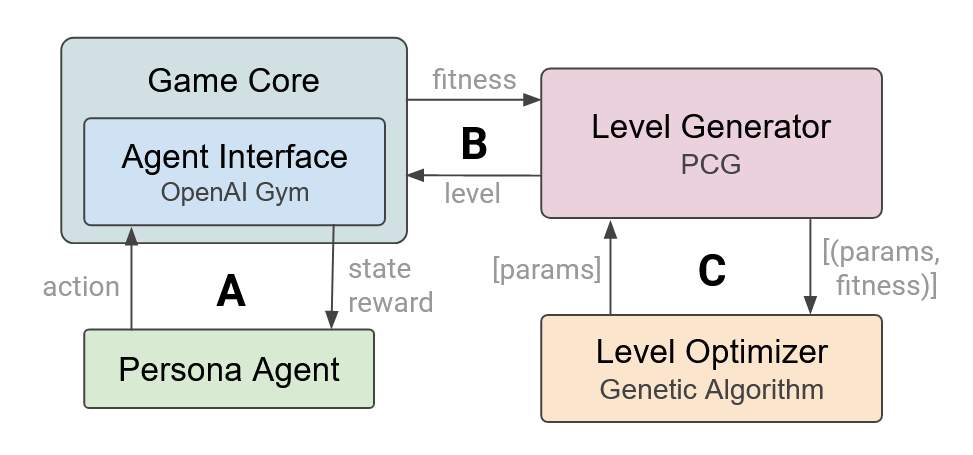}
\caption{The system architecture. The \textbf{A} cycle runs at every game step, the \textbf{B} cycle at the end of every level and the \textbf{C} cycle after a population of levels have been played.}
\centering
\end{figure}

\subsubsection{\textbf{Persona Agent}}



In this paper, we refer to both agents and personas. For the context of this project, their relationship is that an agent \textit{represents} a persona. Therefore, a persona agent is an implementation of the general behaviour exhibited by a given persona. This agent can be of any type, either trained from player data or hand-coded, provided that the agent designer knows about the typical strategies employed in a game.

\subsubsection{\textbf{The Game}}

In theory, any game that satisfies the following constraints would work for this architecture:
\begin{itemize}
    \item Uses a parameterized procedural content generator.
    \item Offers a two-way interface for the game playing agent.
\end{itemize}

Preferably, the game loop should have an unbounded tickrate, so the simulation can run as fast as possible.

For this paper, we used an OpenAI gym-compatible environment \cite{gym} that is capable of using levels produced by a generator. This is not a hard constraint and any other framework which allows for agents to interact with the game/environment could be used, assuming the persona agents are compatible with such framework.

\subsubsection{\textbf{Level Generator}}

A procedural level generator that provides levels that are compatible with the game environment. This generator should have at least one parameter that influences the generated levels. For this project, we use a level generator with 10 numeric parameters (Sec~\ref{sec:generator}), although we also run the experiments using only one of the parameters (Sec.~\ref{sec:res:seed}).

\subsubsection{\textbf{Level Optimizer}}

\label{sec:optimizer}

An optimization algorithm that produces parameters compatible with the level generator. In our experiments this is a simple genetic algorithm where the gene consists of a set of numerical parameters for the generator.

In practice, the optimizer component should provide two functions: one to generate an initial list of level parameters, and the other for generating a new list based on a given input list of parameters, but with a fitness value for each element as well. In theory, any algorithm that is capable of returning meaningful list of parameters with respect to fitness can take part in this architecture.



\section{Implementation}
\label{sec:impl}


In this section, we describe how we implemented the aforementioned architecture to a specific game and set of persona agents. We begin by presenting the game, its mechanics and both the level generator and optimizer, followed by the experience metrics used in our experiments. Finally we outline the behaviour of the four designed persona agents.

\subsection{Game}
\label{sec:game}

"Grave Rave", the game used for this project, is designed to be simple enough for a proof of concept and fast execution, but complex enough so that there are several viable strategies to win. Both the PCG game levels and the enemy behaviour are dependent on input to the level generator. The workings of this level generator is described in Sec.~\ref{sec:generator}. The game is implemented in Python, using pygame\footnote{Library for making games with Python \url{https://www.pygame.org/}}, and the graphics are modified public domain sprites from OpenGameArt\footnote{\url{https://opengameart.org/}}. Both the game and the experiments are hosted on GitHub\footnote{\url{https://github.com/Jontahan/roguelife}}.

Genre-wise, the gameplay is inspired by roguelikes\footnote{Genre of games with elements reminiscent of the game \textit{Rogue} (Michael Toy, Kenneth Arnold, Glenn Wichman, 1980)} and tower defence games\footnote{Games where you protect a base by building "towers" that will shoot at incoming enemy waves}. The player character spawns right next to their base. Five enemies (rendered as skeletons) are spawned at various random places around the map. The enemies begin buried in graves, but after 20 steps, they emerge and start making their way towards the base. When enemies are adjacent to the base, they will attack it and reduce its health. The player wins the game by killing all enemies and loses if the base is destroyed.

\label{implementationEnemyBehaviour}
\textbf{Enemy Behaviour:} After rising from their graves, the enemies will rush towards the base following the shortest Manhattan path. However, when their Euclidean distance to the player becomes less than the \textit{Flee Distance} variable set by the level generator (Sec.\ref{sec:generator}), they will run in the opposite direction of the player. This variable alters the way enemies behave, being more or less "brave" depending on its value. To ensure that the player does not exploit this behaviour by standing next to the base constantly, the enemies will always rush to the base after 70 game steps, irregardless of the players position.

\textbf{Actions:} On every step of the game, both the player and the enemies do a single action. The player can either move or attack, while the enemies can only move (their attack is automatic when adjacent to the base). Movement input is one of the four cardinal directions. Attack input will spawn a projectile on the tile the player is currently facing, which travels into that direction until impact, killing enemies, destroying trees or being stopped by rocks or level boundaries. Trees can also be destroyed by enemies, by being "walked into" thrice.

\subsection{Level Generator}
\label{sec:generator}

To achieve comprehensive level generation, we decided to give the level generator indirect control over the position of all elements in a level and influence over the behaviour of the enemies. It receives $10$ parameters and returns a playable level based on these. The level returned for the same set of parameters will always be the same. Given a set of parameters, a level is generated through the following algorithm:

\begin{enumerate}
  \item A random seed is set as $Random Seed \in [1, 9999]$.
  \item Each position of a $15\times15$ level matrix becomes a rock with a probability given by $Initial Rock Density \in [0.1, 0.4]$. The matrix is then iterated $Rock Refinement Runs \in [1, 3]$ times. At each iteration, every position of the matrix is visited. If the position is not a rock and there are more than $ Rock Neighbour Number \in [4,8]$ neighboring rocks at an Euclidean distance of $Rock Neighbour Depth \in [1,2]$ or less, that position becomes a rock.
  \item The level is tested to see if any free position can be reached from any other free position, which ensures playability. If not, the algorithm returns to step 2.
  \item Each position of the level matrix not occupied by a rock becomes a tree with a probability given by $Initial Tree Density \in [0.1, 0.4]$. The level matrix is then iterated $Tree Refinement Runs \in [1, 3]$ times. If the position is free and there are more than $Tree Neighbour Number \in [4,8]$ neighboring trees at an Euclidean distance of $Tree Neighbour Depth \in [1, 2]$ or less, that position becomes a tree.
  \item A list is populated with the positions that have no neighboring trees or rocks. If the list is empty, the algorithm returns to step 2. Else, a random element of the list becomes the position of the base and a random neighbouring tile becomes the position of the player.
  \item Finally, $5$ enemy graves are randomly positioned on the map on any position that does not have any other object or the player. These skeletons are given a variable, $Flee Distance \in [0,10]$, which will influence their behaviour as explained in Sec.~\ref{implementationEnemyBehaviour}. All enemies in a level are given the same value for this variable.
\end{enumerate}

\begin{figure}[!htb]

\minipage{0.48\linewidth}
  \includegraphics[width=\linewidth]{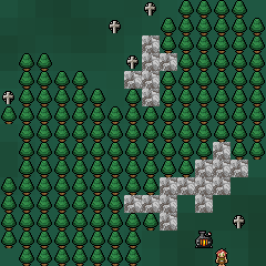}
  \subcaption{}\label{fig:scale_free}
\endminipage\hfill
\minipage{0.48\linewidth}

    \includegraphics[width=\linewidth]{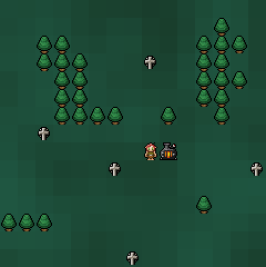}
  \subcaption{}\label{fig:0_99_1cost}
 
\endminipage\hfill

\minipage{0.48\linewidth}

    \includegraphics[width=\linewidth]{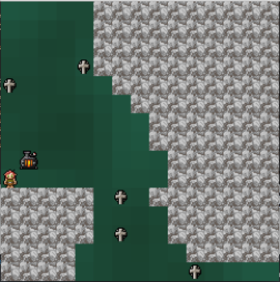}
  \subcaption{}\label{fig:greedy_0_99_1cost}
 
\endminipage\hfill
\minipage{0.48\linewidth}

    \includegraphics[width=\linewidth]{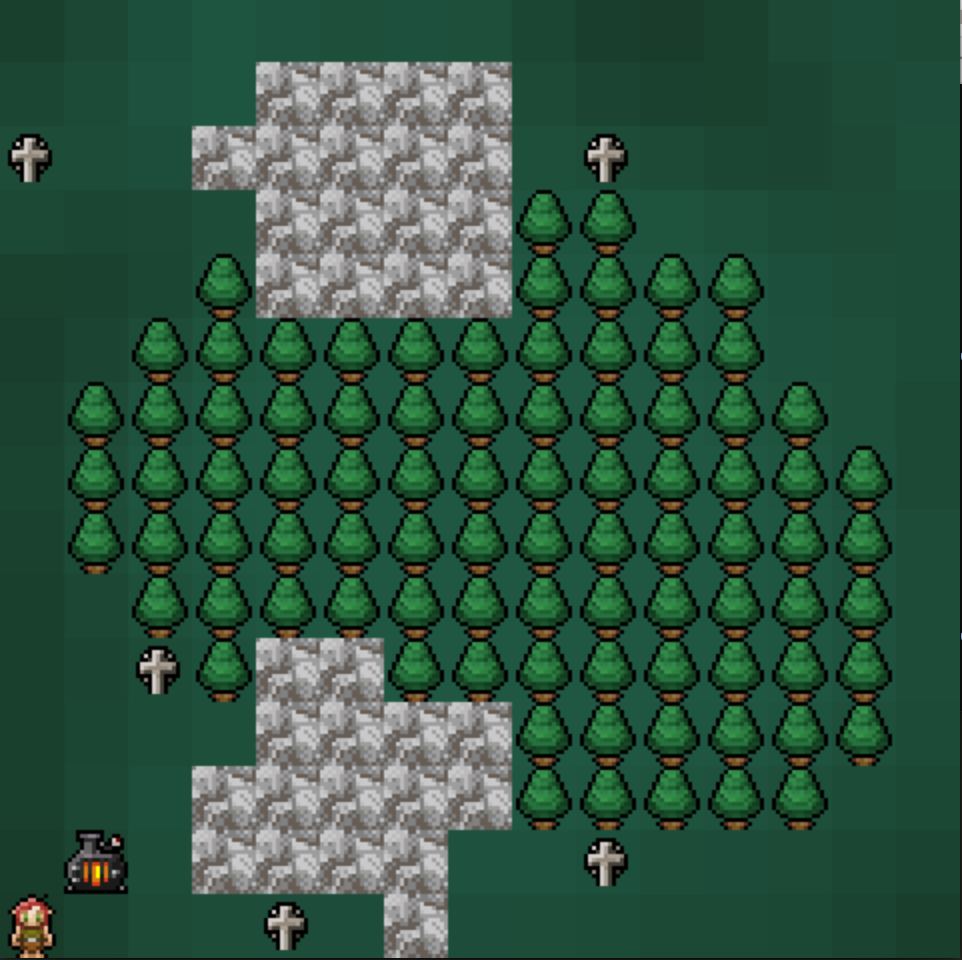}
  \subcaption{}\label{fig:greedy_0_99_1cost}
 
\endminipage\hfill
\caption{Four example levels created by the level generator. The position of the enemies, the position of the base and the density and location of rocks and trees make it so that each of the levels presents a different experience for the player, be it a real player or an agent.}
\label{fig:level_examples}
\end{figure}

Examples of levels generated can be found on Fig.~\ref{fig:level_examples}.

\subsection{Level Optimizer}

For this work and given the relatively low complexity of the game, a simple genetic algorithm was used. Each generation, an agent plays the entire population of levels, assigning each level a fitness given by the experience metric that the optimizer is maximizing (Sec.~\ref{sec:metrics}). Having assigned fitness values to the population, the optimizer selects the top $30\%$ of levels with highest fitness to be propagated to the next generation. All remaining levels are discarded. The selected levels are then repeatedly copied, each of its parameters having a $5\%$ chance of mutation during the copying process. The population for the next generation is thus fully generated when enough copies have been made to reach the original population size. This process is repeated for a set number of generations. 

To explore the flexibility of our architecture, we experimented with using 3 different experience metrics to calculate the fitness of the generated levels.


\subsection{Experience Metrics}
\label{sec:metrics}

For the context of this game, we implement three different experience metrics. Each captures a different experience for a level and play-style pair. These particular metrics focus on challenge, which is often considered a major component of entertainment and overall experience (e.g. Yannakakis and Hallam \cite{yannakakis2007modeling}). Any other metric that could be coded into a numeric fitness function could be used instead of the ones here presented.

For the context of this paper, we define challenge as a player-level property that reflects how poorly a given player agent performs with regard to gameplay objectives.

\subsubsection{\textbf{"Hardcore"}}
\label{sec:optimizer_hardcore}
The "hardcore" metric values maximized challenge. For our game, the agent obtains a reward of $1$ whenever it kills a skeleton. It gets a reward of $-1$ whenever the base loses health. We thus defined the "hardcore" metric as the negated cumulative reward an agent obtained throughout a level. For example, if the agent in a level was able to kill all 5 enemies and the base lost 8 health in the process, the cumulative reward of that agent would be $-3$ (5 enemies killed - 8 base health lost) and the "hardcore" metric would thus be 3 ($-(-3)$).

Based on this metric, we define the fitness function $f_{hc}$ as:
\begin{equation}
    {f_{hc}}(l,a) = -((5 - E) - (10 - B))
\end{equation}
with $l$ being a level, $a$ an agent, $E$ the final number of enemies and $B$ the final health points of the base.

\subsubsection{\textbf{"Easy"}}
\label{sec:optimizer_easy}
The "easy" metric is the opposite of the "hardcore" metric, returning the highest value for levels that present the lowest challenge to the agent. As such, we define the corresponding fitness function, $f_{ez}$, as:
\begin{equation}
    {f_{ez}}(l,a) = -{f_{hc}}(l,a) = ((5 - E) - (10 - B))
\end{equation}
with $l$ being a level, $a$ an agent, $E$ the final number of enemies and $B$ the final health points of the base.

\subsubsection{\textbf{"Close Call"}}
\label{sec:optimizer_close_call}
Besides the extremes of maximum and minimum challenge, we also decided to create a metric that represented the borderline experience, that is, the experience of winning when one is very close to losing. In the context of our game, we interpret this as winning when the base has only one health point left. We call this the "close call" metric, which must therefore be higher when the final base health is lower, unless it becomes zero and the player loses the game, which should correspond to the lowest fitness.
To represent this metric, we defined fitness function $f_{cc}$ as:
\begin{equation}
    f_{cc}(l, a)= 
\begin{cases}
   (10 - B),& \text{if } B \geq 1\\
    -1,              & \text{if } B = 0
\end{cases}
\end{equation}
with $l$ being a level, $a$ an agent and $B$ the final health points of the base.

\subsection{Personas}
\label{sec:personas}

For the purpose of this paper, we designed 4 different rule-based agents able to play the game in different ways, which were lovingly named R01, R02, R03 and R04\footnote{The R is short for "rule-based", and was historically used to distinguish from other agent types in earlier iterations of the project. The two-digit numbering is kept for aesthetics.}. Our objective when designing said agents was to simulate different playing personas, which would react differently to the same level despite all acting to win, that is, to kill all enemies. 

\subsubsection{\textbf{R01}: The Shooter}
As enemies approach the base, the R01 persona will assign each enemy a threat level based on how close to the base they are. They will then walk towards the most threatening enemy using path-finding and shot towards them if they are in line of sight. The agent considers both trees and rocks as obstacles and will not attempt to destroy trees to reach an enemy.
\subsubsection{\textbf{R02}: The Defender}
While the R01 agent will always pursue the closest enemy to the base, the R02 persona will only do so if the enemy is closer than a distance threshold. If all enemies are further away from the base than this threshold, the R02 persona will head back to the base and walk around it until an enemy gets close enough. Other than that, it behaves like R01.
\subsubsection{\textbf{R03}: The Hunter Grave-robber}
The previous personas would always shoot towards enemies when able, regardless if they were still buried under their graves (meaning they are not active and do not pose a direct threat) or not. The R03 persona will instead step on graves while it can, prematurely destroying the enemies. After they rise, the R03 persona behaves exactly like the R01 agent.
\subsubsection{\textbf{R04}: The Mad Man}
All previous personas base the threat level enemies on their distance to the base. The R04 persona uses the distance to itself instead. In other words, this persona will attempt to destroy all enemies, but does not care about the base. It will also step on the closest grave to itself. On top of that, the R04 agent will attack and destroy trees if they stand in the way of the shortest path to the closest enemy.

\section{Results}
\label{sec:results}

\subsection{Baseline}
As a baseline, we ran each agent on 1000 randomly generated levels and collected their episodic rewards. The resulting average reward for each agent can be found on Table~\ref{tab:random_levels}.

\begin{table}[h]
\caption{Average agent reward on 1000 randomly generated levels}
\label{tab:random_levels}
\centering
\begin{tabular}{|c||c|c|c|c|}
\hline
Persona Agent & R01 & R02  & R03 & R04 \\
\hline
Avg. Reward & -0.59 & -0.52 & 0.87 & 0.69\\
\hline
\end{tabular}
\end{table}

The total accumulated episodic reward for this game will always lie in the range $[-10, 5]$, as each enemy kill provides 1 point (there are 5 enemies) and each base health loss provides -1 point (the base has a total health of 10). Given the game rules, a cumulative reward below -4 means the player lost  whereas a reward of -4 or more means the player won.

On average, R03 and R04 appear to fare better than R01 and R02. This is likely because they will step on graves when possible.  



\subsection{The "Hardcore" Metric}

We began our experiments by generating levels that maximized the "Hardcore" metric (Sec.~\ref{sec:optimizer_hardcore}) for each of the previously defined 4 agents (Sec.~\ref{sec:personas}). To do so, we evolved a population of 100 levels over 30 generations for each agent. The evolution algorithm used was the one described in Sec.~\ref{sec:optimizer}, selecting for propagation the levels that were most challenging for the agents (Sec.~\ref{sec:optimizer_hardcore}). We repeated this 15 times to ensure we had consistent results.

\begin{figure}[!htb]
\includegraphics[width=\linewidth]{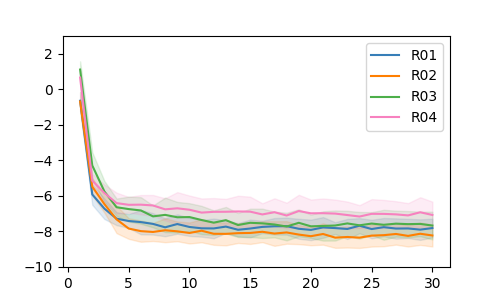}
\caption{Evolution of the average reward for the 4 different agents in the "Hardcore" metric on a population of 100 individuals over 30 generations, showing the standard deviation for 15 runs.}
\label{fig:avg_rewards_hc}
\end{figure}

The evolution of the average reward obtained by the agents on each generation can be seen on Fig.\ref{fig:avg_rewards_hc}. We can observe how the average reward sharply decreases as the generations advance for all agents tested, reaching an equilibrium between -6 and -8. To note that the equilibrium value is slightly different between agents, as is the initial average reward, which is to be expected as some of the agents proved to fare better than others when facing a random level (Tab.~\ref{tab:random_levels}).

In addition to showing how the system converges during generation, we wanted to explore how the agents fared, on average, in levels generated for each other. To do so, we calculated "specificity matrices", where each row represents a persona agent, and each column represents levels that have been adapted to a respective persona. Table \ref{tab:spec_matrix_hc} shows the specificity matrix for the hardcore scenario, and we observe that the cells along the diagonal have the lowest value in the row/column. This makes sense, as the diagonal contains cells where persona agents are tested on levels adapted specifically for themselves, and the hardcore metric should optimize for minimal reward.

\begin{table}[!ht]
\caption{Average agent reward specificity matrix for generated levels for the "Hardcore" scenario, showing the average for 15 runs. The cells in \textbf{bold} are expected to have the lowest reward}
\label{tab:spec_matrix_hc}
\centering
\begin{tabular}{|c||c|c|c|c|}
\hline
& Gen. for R01 & Gen. for R02  & Gen. for R03 & Gen. for R04 \\
\hline \hline
R01 & \textbf{-7.79} & -4.18 & -3.82 & -1.99\\
\hline
R02 & -5.28 & \textbf{-8.21} & -4.82 &  -0.76\\
\hline
R03 & -6.67 & -2.62 & \textbf{-7.66} & -4.10\\
\hline
R04 & -3.82 & -3.20 & -4.15 & \textbf{-7.09}\\
\hline
\end{tabular}
\end{table}

\subsection{The "Easy" Metric}

\begin{figure}[!htb]
\includegraphics[width=\linewidth]{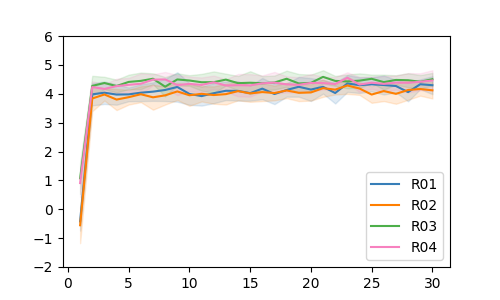}
\caption{Evolution of the average reward for the 4 different agents for the "Easy" metric on a population of 100 individuals over 30 generations, showing the average with standard deviation for 15 runs.}
\label{fig:avg_rewards_ez}
\end{figure}

Next, we decided to run the exact same experiment but this time maximizing the "Easy" metric. The resulting evolution of cumulative reward for the 4 agents over 30 generations can be found on Fig.~\ref{fig:avg_rewards_ez}. We see the opposite reward evolution that we did on Fig.~\ref{fig:avg_rewards_hc}, having the average reward increase until it reaches an equilibrium roughly between 4 and 5.

We further generated the specificity matrix for this metric, which can be seen on Table~\ref{tab:spec_matrix_ez}.

\begin{table}[!ht]
\caption{Average reward specificity matrix for the "Easy" scenario, showing the average for 15 runs. The cells in \textbf{bold} are expected to have the highest reward}
\label{tab:spec_matrix_ez}
\centering
\begin{tabular}{|c||c|c|c|c|}
\hline
& Gen. for R01 & Gen. for R02  & Gen. for R03 & Gen. for R04 \\
\hline \hline
R01 & \textbf{4.30} & 1.68 & 1.82 & 0.65\\
\hline
R02 & 0.04 & \textbf{4.12} & -0.13 &  0.24\\
\hline
R03 & 3.64 & 3.25 & \textbf{4.50} & 2.48\\
\hline
R04 & 1.99 & 3.24 & 3.01 & \textbf{4.46}\\
\hline
\end{tabular}
\end{table}

\subsection{The "Close Call" Metric}

\begin{figure}[!htb]
\includegraphics[width=\linewidth]{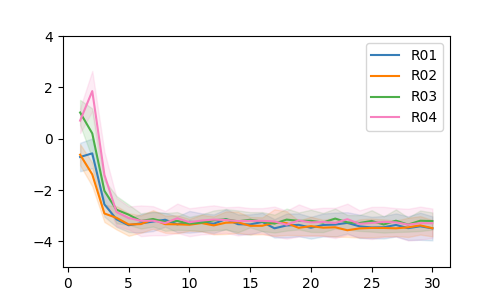}
\caption{Evolution of the average reward for the 4 different agents for the "Close Call" metric on a population of 100 individuals over 30 generations, showing the average with standard deviation for 15 runs.}
\label{fig:avg_rewards_cc}
\end{figure}

We also ran the same experience maximizing the "Close Call" metric. The resulting evolution of reward can be seen on Fig.~\ref{fig:avg_rewards_cc}, showing that the average reward for all agents reaches an equilibrium roughly between $-3$ and $-4$. As previously mentioned, a value of cumulative reward below $-4$ means the player lost the game. As such, an equilibrium slightly above $-4$ means that the levels being generated are, on average, near as challenging as possible while still allowing the player to win.

The specificity matrix for the "Close Call" metric can be found on Table~\ref{tab:spec_matrix_cc}. Contrary to the specificity matrices for the previous two metrics, the diagonal doesn't show perfect specificity. On average, the R01 persona agent playing the levels generated for R03 gets a cumulative reward closer to $-4$ than the R03 persona agent. Upon further inspecting the results, we found that this closeness to the $-4$ value was a consequence of being an average for 15 runs, with the reward for each run ranging all the way from $-7.5$ to $3.31$. In comparison, the runs for R03 playing the levels generated for R03 ranged only from $-3.57$ to $-2.17$.

We thus decided to also generate the specificity matrix for the final base health instead of reward, which can be found in Table~\ref{tab:spec_matrix_cc_hp}. In this matrix, we can already see specificity, although not as well defined as for the previous metrics. The high variance between runs for agents playing levels evolved for other agents makes specificity matrices made for non-borderline metrics, like the "easy" and "hardcore", harder to interpret given the effects of averaging. Closely examining the results, however, allows us to see the specificity is still there, as agents playing levels not evolved for them have highly variable values of reward and final base health.


\begin{table}[h]
\caption{Average reward specificity matrix for the "Close Call" scenario, showing the average for 15 runs. The cells in \textbf{bold} are expected to have a reward close to -4}
\label{tab:spec_matrix_cc}
\centering
\begin{tabular}{|c||c|c|c|c|}
\hline
& Gen. for R01 & Gen. for R02  & Gen. for R03 & Gen. for R04 \\
\hline \hline
R01 & \textbf{-3.50} & -2.30 & -3.97 & -1.42\\
\hline
R02 & -1.21 & \textbf{-3.48} & -2.93 &  -2.01\\
\hline
R03 & -1.60 & -0.55 & \textbf{-3.21} & -1.04\\
\hline
R04 & 0.14 & -2.06 & -3.09 & \textbf{-3.30}\\
\hline
\end{tabular}
\end{table}

\begin{table}[h]
\caption{Average base hp specificity matrix for the "Close Call" scenario. The cells in \textbf{bold} are expected to have a value close to 1, but not below}
\label{tab:spec_matrix_cc_hp}
\centering
\begin{tabular}{|c||c|c|c|c|}
\hline
& Gen. for R01 & Gen. for R02  & Gen. for R03 & Gen. for R04 \\
\hline \hline
R01 & \textbf{1.64} & 3.69 & 2.16 & 4.42 \\
\hline
R02 & 4.31 & \textbf{1.67} & 2.63 & 3.68 \\
\hline
R03 & 4.16 & 5.06 & \textbf{1.85} & 4.32\\
\hline
R04 & 5.69 & 3.83 & 2.56 & \textbf{1.83}\\
\hline
\end{tabular}
\end{table}

\subsection{Random Seed Only}
\label{sec:res:seed}

\begin{table}[!ht]
\caption{Average reward specificity matrix for the "Hardcore" metric evolving only the $Random Seed$ parameter, showing the average for 15 runs. The cells in \textbf{bold} are expected to have the lowest reward.}
\label{tab:spec_matrix_hc_s}
\centering
\begin{tabular}{|c||c|c|c|c|}
\hline
& Gen. for R01 & Gen. for R02  & Gen. for R03 & Gen. for R04 \\
\hline \hline
R01 & \textbf{-7.74} & -2.92 & -5.73 & -3.47\\
\hline
R02 & -3.62 & \textbf{-8.20} & -3.91 &  -2.75\\
\hline
R03 & -3.64 & -3.97 & \textbf{-7.71} & -1.97\\
\hline
R04 & -1.91 & -4.35 & -0.73 & \textbf{-7.41}\\
\hline
\end{tabular}
\end{table}

Finally, after obtaining the previously described results, we wondered if our architecture would still work if we evolved only the $Random Seed$ parameter. Games like Minecraft, Terraria, and many others, generate their levels based on an initial random seed and no further parameters. If we could obtain similar results to the ones we previously obtained while only evolving the random seed, we could expect our architecture to be applicable to a greater number of PCG generated games without forcing them to undergo any significant modifications.

We therefore decided to repeat the previous experiments, but this time, we only allowed the $Random Seed$ parameter to undergo evolution. The other parameters were randomly assigned when generating the initial population. To compensate the loss of mutable parameters, the $Random Seed$ parameter had a $20\%$ chance of mutation whenever it was copied.

\begin{figure}[!htb]
\includegraphics[width=\linewidth]{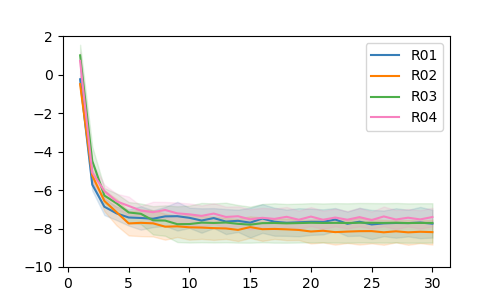}
\caption{Evolution using only the $Random Seed$ parameter of the average reward for the 4 different agents in the "Hardcore" scenario on a population of 100 individuals over 30 generations, showing the standard deviation for 15 runs.}
\label{fig:avg_rewards_hc_s}
\end{figure}

We found that, for the "Grave Rave" game, evolving only the seed led to comparable results, over all metrics, to evolving all the parameters.

As an example, the evolution of the average reward for the 4 different agents for the $Random Seed$ only evolution, maximizing the "Hardcore" metric, can be seen in Fig.~\ref{fig:avg_rewards_hc_s}. It shows a comparable rate of convergence and final average reward to the results obtained when evolving all the parameters (Fig.~\ref{fig:avg_rewards_hc}). The specificity matrix (Table~\ref{tab:spec_matrix_hc_s}) also shows a similar level of specificity than the one where all parameters were evolved (Table~\ref{tab:spec_matrix_hc}).

\section{Discussion}
\label{sec:discussion}
\subsection{Personal Adaptation}
For every explored scenario, we observe that the system converges on a satisfactory solution for the persona (Fig~\ref{fig:avg_rewards_hc} to \ref{fig:avg_rewards_hc_s}). A more interesting observation, however, is that, on average, the level adaptation is quite persona specific, as the intended scenario is never maximized to the same extent for the other personas as it is for the persona it is intended for (Table~\ref{tab:spec_matrix_hc} to \ref{tab:spec_matrix_hc_s}). This phenomenon is not by design, but appears as an emergent property of the game itself. For another game, it might be the case that a level that maximizes the "hardcore" metric for one persona might have a very high likelihood of also doing so for many others. It will also heavily depend on how the personas are defined and how different they are from one another.

In an applied context, a designer could tweak our architecture to either ensure a level maximizes a given metric specifically for a single persona or generally over many different personas. One way to ensure either specificity or generality would be to run several evolutionary runs and generate the specificity matrix for the relevant metric for all different runs. The designer could then choose the level populations that had the specificity matrix that better fit the design goals.


\subsection{Scenarios}
In this paper we present three gameplay scenarios: "hardcore", "close call" and "easy". All of these can be found in most games with a win-condition. The results show that the system consistently converges on satisfying levels for each scenario using more or less the same number of generations. As the initial experiments converged relatively fast in wall-clock time, little effort was put into runtime optimization in this project. However, if necessary, we would apply more sophisticated methods from evolutionary algorithms.

One flaw of this system for an applied context, is that designing gameplay scenario metrics can be difficult for certain games. While envisioning a scenario to represent the "ideal player experience" is a natural exercise for the game designer, implementing a numerical distance metric to this scenario is far from trivial. In addition, it is naive to assume that a game has only one such scenario, which suggests that an appropriate function for the metric could be a complex one, with branching and dependencies on many different types of information from the game, all of different scales and formats.

\subsection{Seed-only Evolution}

For the "Grave Rave" game, we found that evolving only the $Random Seed$ parameter led to similar results to evolving all 10 parameters. This is a promising result for the generality of the approach. However, with the seed being functionally discrete, the only way it can be used in a mutation context is to either copy it or re-initialize. This disables many potential improvements to the genetic algorithm, such as sexual reproduction. Without specific investigation, we can only guess that our search problem is a rather simple one, seeing as most scenarios appear to converge after roughly 10 generations. For more complex PCG games, intuition suggests that making smaller adjustments to the level might allow for a more stable search.
\section{Conclusion}
\label{sec:conclusion}
This paper introduces a specific framework for evolving levels to player personas. The framework is technically compatible with any agent type that can interface towards OpenAI gym, and any game that takes numerical parameters to their PCG, and can provide meaningful metrics for measuring player experiences. One of the main objectives of this project is to take ideas that have been around for a while already, and bring them closer to commercial application. 

Three distinct experience metrics were tested, each returning successful results for all four persona agents. In addition to consistent success in finding levels that satisfy the desired scenario, we also observe that the system adapts to the given persona specifically, instead of providing general persona-independent levels. While our initial experimentation shows promise, there are likely many obstacles for this to scale to commercial games, both expected and unexpected. Some of these obstacles include training time, persona representation and designing experience metrics. 

The choice of hand-coded persona agents is sufficient for demonstration purposes, but have no data basis from collected player behaviour. The purpose is to show how the architecture would function if given a set of distinct persona agents. The experience metrics are more intentionally designed by humans, as a game designer would likely have a hand in defining what the "ideal experience" looks like. However, for future work, the metrics could be derived from data as well, such as biometric stimuli.




\section{Future Work}
\label{sec:future_work}

Given the results of this project, one of the more natural follow-up directions is to explore adaptation specificity versus generality. In other words, testing whether adapting to all persona agents (general adaptation), is a compromise on how well the resulting adaptation is with respect to the gameplay scenario metric. This includes experiments where each level is evaluated on all persona agents, and combining the fitness produced by each, either as an average (for general adaptation), or penalizing for all but one persona (specific adaptation).

Furthermore, as this paper is limited to one single, homemade game, it is important to assess whether similar results can arise from other, preferably commercial, titles. One major roadblock for this is that our system requires access to a parametrized content generator. The seed-only approach tested in Section \ref{sec:res:seed} enables considerably more games to be used, but we have reasons to doubt how well this will scale, as evolving only one discrete value restricts how much we can take advantage of most optimization techniques developed for evolutionary algorithms.

Using rule-based persona agents has the advantages of having predictable and consistent behaviours as well as requiring no recorded player data to create. However, this comes at the cost of only representing strategies that the agent designer comes up with, potentially leaving out highly relevant approaches to the game. There is also no assurance such agents truly represent any real player personas. In addition, designing several diversely behaved bots by hand for a game might be close to impossible for many complex games. Alternative approaches to acquiring persona agents include using RL agents with persona specific rewards, learning strategies from player data \cite{pm-persona}, or using procedural personas \cite{automated-playtesting}.

\section*{Acknowledgments}
The authors would like to thank the people behind the 2021 edition of the Summer School of AI and Games, where the authors got to know each other. We would also like to thank Even Klemsdal for his contributions in the earlier stages of the project, and the Media Technology program at Leiden University for their support.
This work was supported by national funds through Fundação para a Ciência e a Tecnologia (FCT)
with reference UIDB/50021/2020, by grant 2020.05865.BD and by the EU H2020 RIA project iv4XR : 856716.

\bibliographystyle{IEEEtran}
\bibliography{references}
\end{document}